\documentclass{IOS-Book-Article}
\usepackage{graphicx} 
\usepackage{amsmath}
\usepackage{xspace}
\usepackage[nodisplayskipstretch]{setspace}
\usepackage{todonotes}

\newcommand{\part}{\textsc{Part}\xspace}

\newcommand{\containment}{\textsc{Containment}\xspace}

\newcommand{\movement}{\textsc{Movement}\xspace}

\newcommand{\mover}{\textsc{Mover}\xspace}

\newcommand{\sourcepathgoal}{\textsc{Source\_Path\_\allowbreak{}Goal}\xspace}

\newcommand{\support}{\textsc{Support}\xspace}
\newcommand{\supporter}{\textsc{Supporter}\xspace}
\newcommand{\supported}{\textsc{Supported}\xspace}
\newcommand{\contact}{\textsc{Contact}\xspace}
\newcommand{\transportation}{\textsc{Transportation}\xspace}

\newcommand{\ISL}{\textsf{ISL}$^{FOL}$\xspace}

\begin{document}
\begin{frontmatter}
%
\title{Hanging Around: Cognitive Inspired Reasoning for Reactive Robotics}
\runningtitle{FOIS RT}

%
\author[A]{\fnms{Mihai} \snm{Pomarlan}},
\author[B]{\fnms{Stefano} \snm{De Giorgis}},
\author[C]{\fnms{Rachel} \snm{Ringe}},
\author[D]{\fnms{Maria M.} \snm{Hedblom}},
\author[E]{\fnms{Nikolaos} \snm{Tsiogkas}}
\address[A]{\textit{Applied Linguistics Department} \textit{University of Bremen}, Bremen, Germany}
\address[B]{\textit{Institute of Cognitive Sciences and Technologies} \textit{National Research Council}, Catania, Italy}
\address[C]{\textit{Digital Media Lab} \textit{University of Bremen}, Bremen, Germany}
\address[D]{\textit{Department of Computing} \textit{J{\"o}nk{\"o}ping School of Engineering}, J{\"o}nk{\"o}ping, Sweden}
\address[E]{\textit{Department of Computer Science} \textit{KU Leuven}, Leuven, Belgium }
\begin{abstract}
    Situationally-aware artificial agents operating with competence
    in natural environments face several challenges: spatial awareness, object affordance detection, dynamic changes and unpredictability.
    A critical challenge is the agent's ability to identify and monitor environmental elements pertinent to its objectives. Our research introduces a neurosymbolic modular architecture for reactive robotics. Our system combines a neural component performing object recognition over the environment and image processing techniques such as optical flow, with symbolic representation and reasoning. The reasoning system is grounded in the embodied cognition paradigm, via integrating image schematic knowledge in an ontological structure. The ontology is operatively used to create queries for the perception system, decide on actions, and infer entities' capabilities derived from perceptual data. 
    The combination of reasoning and image processing allows the agent to focus its perception for normal operation as well as discover new concepts for parts of objects involved in particular interactions. The discovered concepts allow the robot to autonomously acquire training data and adjust its subsymbolic perception to recognize the parts, as well as making planning for more complex tasks feasible by focusing search on those relevant object parts. We demonstrate our approach in a simulated world, in which an agent learns to recognize parts of objects involved in support relations. While the agent has no concept of handle initially, by observing examples of supported objects hanging from a hook it learns to recognize the parts involved in establishing support and becomes able to plan the establishment/destruction of the support relation. 
    This underscores the agent's capability to expand its knowledge through observation in a systematic way, and illustrates the potential of combining deep reasoning with reactive robotics in dynamic settings.
\end{abstract}


\end{frontmatter}


\section{Introduction}
\label{sec:intro}

A complex tapestry of latent knowledge underpins every interaction between an agent and its environment. 
Aspect such as affordances, agent's own capabilities, and nuanced properties of the environment form the bedrock upon which cognitive agents perceive, interpret, and navigate their surroundings.
The depth of this interaction is not merely a function of the agent's immediate sensory input but is influenced by a pre-existing, albeit latent, framework of knowledge. This framework includes stored interaction patterns and hard-wired relations that dictate the agent's engagement with its environment, guided by the rules that govern the world in which the agent is operating.

Despite impressive advancements in generative AI ``foundational models,'' a significant gap remains in their understanding and representation of physical and spatial dynamics.
Most recent OpenAI release, SORA\footnote{SORA's technical report is available here:\newline \url{https://openai.com/research/video-generation-models-as-world-simulators}}, is a Language and Vision Model to generate video from text prompts.
While the generated contents at this time are very good looking, they also showcase limitations such as a poor grasp of object permanence -- entities may flick in and out of existence in implausible ways -- 
and elementary physics laws~\cite{GaryMarcus2024substackSORA}.


Endorser of the ``more data is all you need'' claim that such errors will eventually be fixed, and this is not, a-priori, a vain hope: even large neural networks have to ``compress'' its training data, and in so doing, they will stumble upon regularities of the world.
We are reluctant to endorse this view however. Generative AI models, while groundbreaking in generating coherent and contextually relevant linguistic or visual outputs, primarily operate on the basis of the statistical probability of sentence or image completion, with statistics obtained from a corpus of decontextualized recordings. This approach has a fundamental limitation: the lack of embodied grounding, informed by real-world sensorial data, obtained and interpreted by an agent in purposeful engagement with that world. 

More than ever, SORA's impressive results raise questions like:
what is that world knowledge we need, how could it be described, how would it be used in an autonomous agent interacting with some kind of world?

Furthermore, SORA and LLM's lack of embodied grounded knowledge is an echo of
Moravec's paradox~\cite{moravec1988mind}: the observation that human intuitions for what is cognitively easy do not translate to machines.
The problem of endowing practical know-how to artificial agents is of chief relevance in autonomous robotics, and it is from this perspective -- that of artificial agents -- that we approach the problem. Thus, though 
our system is cognitively inspired, we do not endeavour to obtain cognitive plausibility.


With the purpose to display how cognitive robotics, and in particular a neuro-symbolic architecture, is capable to perform commonsense reasoning on the world thanks to embodied cognition knowledge, our exploration in this work is driven by the following questions: which
objects exist, where to place attention, and how may an agent enrich its knowledge, at different levels of abstraction, about the entities in the world and their interactions.

Our contribution is threefold: (i) we represent, in a formal way, a semantic parsing of sensorimotor events that a cognitive system undergoes as it observes and interacts with an environment. This event segmentation is grounded in the paradigm of embodied cognition, specifically the notion of Image Schemas, as detailed in Section \ref{sec:cognitive_grounding}.

Second (ii), we propose the development of a modular reasoning system to identify specific situations, the entities involved, and the roles they play. This system is designed to operate at the intersection of neuro-symbolic processing, leveraging the continuous stream of perceptual data obtained from a neural architecture. The fusion of neural inputs with symbolic, ontology-based reasoning allows for the transformation of raw sensory signals into structured, actionable knowledge. This integration
enables to dynamically update the system's internal knowledge to new information and environmental changes.

Lastly, (iii) we investigate how the combination of neural network-based object recognition and reasoning can enlarge
an artificial agent’s ontology. Our agent starts with a certain knowledge base: it knows a set of objects, understood as finite shapes. However, it can teach itself to recognize parts of these objects if those parts are involved in functional relationships, i.e. relationships between objects that constrain how a scene will unfold.

Thus, we introduce a ``sense-making'' system, dedicated to the identification and understanding of mereological affordances within the environment. This system, through the repeated observation of spatio-relational patterns, is adept at recognizing areas of interest that exhibit new \emph{emergent properties}. These properties are not static but evolve based on the functional parts of the environment, thereby presenting a continuously shifting landscape of interaction possibilities.\footnote{Our source code and knowledge modelling is available at https://github.com/heideggerian-ai-v5/fois2024} 


Finally, from an ontological point of view, our work relies on the Description \& Situation (DnS) pattern \cite{gangemi2003understanding,gangemi2008norms}, based on the reification of intensional/extensional relations with recursive accessibility.


\section{Related Work: Conceptual Modeling, Embodied Grounding, and Cognitive Robotics}
\label{sec:cognitive_grounding}

In this section we will briefly review some conceptual tools and existing approaches we use to build our system. We start with a summary on image schemas and their hypothesized roles in human cognition, continue with frame semantics as employed to understand descriptions of world states, and end with a discussion on how image schemas have been previously formalized. Finally we provide references to previous approaches in robotics relying on cognitive paradigms.

\paragraph{\textbf{Image Schemas}}
To have an understanding of the space of possibilities for an artificial agent, it helps to look at what humans seem to do. 
The learning process in children, known as perceptual meaning analysis (PMA) \cite{Mandler2004foundations}, involves deriving generalized spatiotemporal patterns, called image schemas \cite{Johnson87}, from such interactions. Image schemas represent a finite set of relationships among objects, agents, and their environments that define their uses and the spaces of their affordances. Examples are \containment, meaning one object can be/is contained inside another one and \sourcepathgoal, meaning objects can/do move along particular trajectories.

%
%

Image schemas are considered the foundation for reasoning \cite{lakoff1990invariance}, and were shown to evolve into cognitive functions such as natural language and conceptualizations of abstract entities \cite{Johnson87,lakoff2000mathematics} through grounded, experiential patterns. 
Furthermore, image schemas can combine in more complex structures.
Take for instance the notion of ``transportation''. It does not rely on any object in particular, but can be generally understood as the ``movement of object(s) from A to B''. In image-schematic terms, it can be described as a combination of \sourcepathgoal and \support or \containment~\cite{kuhn2007image}. 
By combining these concepts in constellations and sequences (as state spaces)~\cite{hedblom2019image,conf/stAmant06}, it is possible to formally describe the structure of increasingly complex events.

\paragraph{\textbf{Frame Semantics and Its Ontological Modelling}}
For the conceptual modeling we rely on the Frame Semantics cognitive paradigm and reuse the notion of ``frame'', as in Minsky and Fillmore \cite{minsky1974framework,fillmore1982framsemantics}. Frames are schematic abstract representations of recurrent situations. Each frame takes a set of semantic roles, namely the elements which participates to the frame situation. The minimum set of roles to realise a frame composes the ``necessary roles''.
Frames are formalised as N-ary relations with a central node being the Event/Situation, and a number N of semantic roles participating to it. 
Note that the set of possible roles is much broader than the one of its necessary roles. 

From an ontological modeling perspective, employing frames for representations adheres to good modeling practices through the adoption of the Description and Situation \cite{gangemi2003understanding} Ontology Design Pattern. In this framework, each image schema discussed in our study is conceptualized as a semantic frame, represented as a \emph{Description} that is satisfied by a \emph{Situation}. In more detail, the \emph{Situation} reflects a specific world state, encapsulating the essential roles required for its realization. If the \emph{Situation} presents the necessary roles as formalized by a particular \emph{Description}, then the \emph{Situation} \emph{satisfies} the \emph{Description}.
DnS as a formalization of Frame Semantics has been largely used in various projects (mainly in its OWL formalization) \cite{scherp2009f,eschenbach2008formal,hoffner2022deep,beretta2021challenge}.



\paragraph{\textbf{Image Schema Ontological Modeling}}
Formally representing image schemas is a complex problem as they are conceived and treated as abstract ``gestaltic entities'' \cite{Johnson87} without clear borders or structure. However, traditional methods in spatiotemporal reasoning have been proposed as representation approaches (e.g.~\cite{kuhn2007image}) and some of these logical languages and calculi, in particular Region Connection Calculus \cite{randell1992spatial}, Qualitative Trajectory Calculus \cite{van2006qualitative} and Linear Temporal Logic~\cite{LTL2008} were combined into the modelling language the Image Schema Logic, \ISL ~\cite{hedblom2020springer,hedblometal2017}.
Being a description language in first-order logic, \ISL can capture detailed spatiotemporal interactions and transformations that can be used to represent situations or events.

Complex events can be seen as compositions and co-occurrences of more elementary situations or `scenes', which in turn can be represented using image-schematic representations~\cite{hedblom2019image}. In this way, it is possible to decompose events and even robotic action plans based on the sensorimotor input about a situation, exploiting flowing data from a perception module  (see examples in~\cite{conf/stAmant06,hedblom2019image,hedblompomarlan2021dynamic,dhanabalachandran2021cutting}). 

However, for most autonomous systems, this level of spatiotemporal modelling is too detailed to be used for actual real time situation analysis and decision making tasks. Therefore, the fundamental image-schematic representations have been proposed by transposing the methods into different types of description logics, e.g. EL++~\cite{hedblompomarlan2021dynamic} and OWL2\footnote{See the full ISL2OWL graphs at \url{https://github.com/StenDoipanni/ISAAC/tree/main/ISL2OWL}}~\cite{de2022imageschemanet}). In~\cite{Pomarlanetal2022}, we introduced the image-schematic reasoning layer (ISRL) which is based on ISL2OWL, a simple ontological module of image-schematic components.

In ISL2OWL, acting as an ontological module, each image schema is modeled as a semantic frame. More precisely as an N-ary relation with central node the image schematic situation, where the spatial primitives form the necessary roles. For example, a \support situation takes as necessary roles two elements: a \supporter, and a \supported entity. The underlying theoretical assumption is derived from image schematic literature, and directly dependent on the Gestalt~\cite{verstegen2000gestalt}, frame-based nature of image-schemas~\cite{lakoff1980metaphors}: if one of its roles (spatial primitives) is instantiated, this implies the activation of the whole image schema. This means that knowing there is a supporting entity also means knowing there is a supported one, and situation is a Support situation.

Therefore, given a certain state of the world, if a certain entity is retrieved as being in movement, that particular state of the world will be represented as a \movement situation, taking as participant the moving entity as \mover. The same situation could, of course, be qualified by more than one image schematic relation, in a combinatorial increasing degree of complexity.

Furthermore, following ~\cite{hedblom2019image}, there are three possible image schematic combinations: Merge, Collection, and Structure. Thanks to the frame approach, more complex scenarios can be modeled as N-ary relations taking as roles more simple situations, for example, a \transportation situation results from the co-occurrence of a \movement situation co-located with a \support situation by being axiomatised as taking as roles some \movement and \support. Thanks to the reasoning system described in the following, a \movement situation $S1$ having as participants $x$ and $y$, and a \support situation $S2$, taking as participants the same $x$ and $y$, is inferred as \transportation situation.

While a lot of information is abstracted away from the original \ISL, representing the image schemas in a computationally feasible way in ISL2OWL allows for them to be used in logical reasoners and as a consequence, we can represent them as semantic building components for the task descriptors.

\paragraph{\textbf{Cognitive Robotics}}
Allowing a robot to display an intelligent behaviour is the main goal of the field of cognitive robotics. It involves studying the knowledge representation and reasoning problems a robot faces in a dynamic and partially observable world \cite{levesque2008cognitive}. In addition to representing knowledge and reasoning, cognitive robotics studies methods of learning through interaction with the environment \cite{moldovan2012learning, lara2018embodied}.

For a cognitive robot to function in the dynamic and uncertain environment that the real world is, three main components are required: i) a source of knowledge regarding the environment, ii) a computational framework to process this knowledge, and iii) a world representation that models the environment and the robot's behaviours. A combination of these components is named a \emph{Semantic Reasoning Framework} (SRF) \cite{liu2023survey}, from which we inherit the terminology to describe the architecture in the next section.

\section{Semantic Reasoning Framework}
\label{sec:architecture}

\begin{figure}[ht]
\centering
\includegraphics[width=\columnwidth]{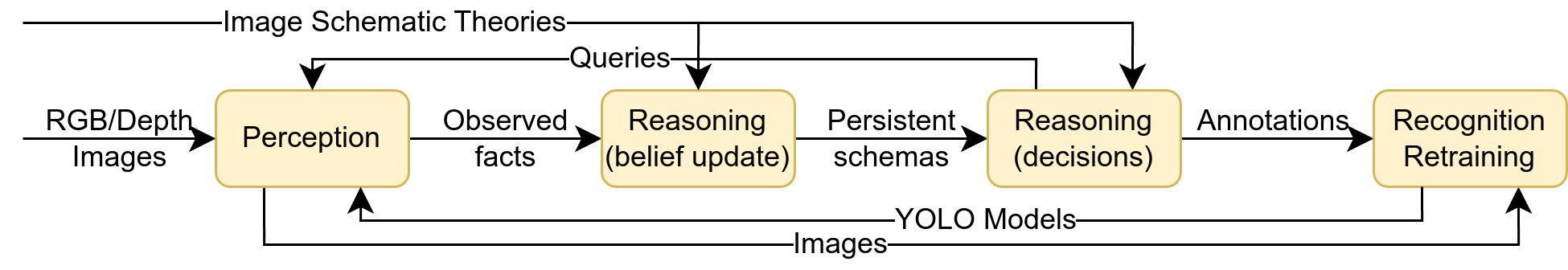}
\caption{Architectural overview of the agent.}
\label{fig:cog_rob_workflow}
\end{figure}

In this section we describe the overarching goal of our agent and its modular structure. The fundamental goal of the agent is to gather information from the environment and interpret it in image schematic terms, so as to maintain an ongoing model of what it observes and engages with, and produce decisions on how to continue that engagement.
A visual representation of the architecture can be seen in Fig. \ref{fig:cog_rob_workflow}. We provide an informal, high-level descriptions of the various modules in the following subsections, to delineate what roles symbolic inferences ultimately play in our approach. We then describe the theories employed for the reasoning task in more formal detail in Section~\ref{sec:agents-theories}.
%

\subsection{Towards Engagement with the World}

A naive understanding of perception would be that, modulo errors that in principle can be eliminated, it constructs a truthful picture of the world out of facts independent of contextual factors such as the goals of the perceiver. Following Heidegger, AI critic Hubert Dreyfus argued against this view and that it is responsible for the stalling of early AI efforts~\cite{Dreyfus92}. In an attempt to simplify and translate a part of Dreyfus' critique in more engineering friendly terms\footnote{The coauthors with a robotics background find such ``translations'' necessary because it is otherwise hard from Dreyfus' philosophy to infer what to actually do in the context of AI research. Dreyfus may have intended to say AI is doomed, but as AI researchers we have to play that game anyway.}, we would say that the fact of noise requires filtering, and filtering requires assumptions as to what is noise and what is meaningful. Thus, an agent is not a passive receiver of facts from the environment via a perception pipeline, i.e. a computational structure that feeds information in only one way.

Instead, an agent must actively choose what and how to look at, based on its current understanding of how it is embedded in a situation. This understanding includes beliefs about what the situation is and the agent's place in it, and how these may change in the future. Perception is then ``taskable'': reconfigurable based on the agent's needs, in order to answer specific queries dictated by aspects such as what the agent expects of a situation. 

Thus, one of our goals in implementing our system, is to investigate the knowledge structures which, if appropriately connected to sensoric and motor procedures, would enable an agent to have an understanding of a situation and its (plausible) evolution. For the sake of exposition we will use a shorthand and speak of a symbolic layer implementing reasoning on image schematic assertions, such as that some object supports another. It should be understood that the meaning of such a statement is not captured merely at the symbolic layer, but rather in how the symbolic inferences rewire sensors and actuators. ``Support'' is just a name, it gets its meaning from what the agent expects to observe and may decide to do and what outcomes this has in the world, including in the agent's own disposition about what to perceive and how to act.

Dreyfus' critique was arguably influential in the development of the related fields of reactive, embodied and situated robotics, a field which he himself later reviewed and described as ``Heideggerian AI''~\cite{Dreyfus2007HeideggerianAI}. We place our work in the field of reactive robotics too, so it is pertinent to notice that Dreyfus' verdict was this AI project also stalled -- a conclusion he would probably maintain today as well.

Again simplifying and ``translating'', Dreyfus observes that reactive robotic systems are nonetheless trapped by their ontology. Reactive robots have a given, finite, inventory of concepts and no ability to produce new ones regardless of any interactions with the world they may experience. Simplifying even more, Dreyfus' challenge is to have an agent able to teach itself to see new things.

To address this challenge we endow our system with the ability to store snapshots of sensor data and automatically annotate them as exemplars of concepts created at the image schematic layer. These ``new concepts'' have a given structure -- ``an object that can play a particular role in a situation satisfying some description'' -- but can in principle be arbitrarily complex based on how intricate the role and situation descriptions are. Simply creating a new concept expression is of course not enough to produce something meaningful, which is why the stored exemplars are used to retrain perception -- literally, training it to see new patterns. Thus, the new concepts become grounded in new perception procedures to recognize them in the world, and in ways to use the new objects once discovered -- the concepts describe what roles they can play.

This is made possible precisely by the interplay of a symbolic layer maintaining an agent's understanding of its situation, and its subsymbolic sensorimotor apparatus. The sensors can partition the world in an infinity of ways -- in other words, the pixels of an image can be clumped arbitrarily -- while the image schematic understanding picks out which ways may be meaningful. This allows the system to somewhat escape the ontology trap seen by Dreyfus, because it is not limited to an initial set of objects it can recognize and which it is forced to treat as atomic. Instead, it can discover ``functional parts'', i.e. partition objects based on how the objects play the roles they play in a situation.

\subsection{The Perception Module}
\label{sec:perception_module}

The perception module offers the agent a conversion between numeric sensor data obtained through an RGB and depth camera and qualitative descriptions. It fulfils this task by answering queries such as object, relative movement, and contact detection. A query is represented as a triple of form $(p, s, o)$ where $p$ indicates the type of query, and $s$ and $o$ are objects. Note that $o$ can also be left blank, in which case the query is interpreted as asking about all objects that move relative to/contact object $s$.

\begin{figure}[ht]
\centering
\includegraphics[width=\columnwidth]{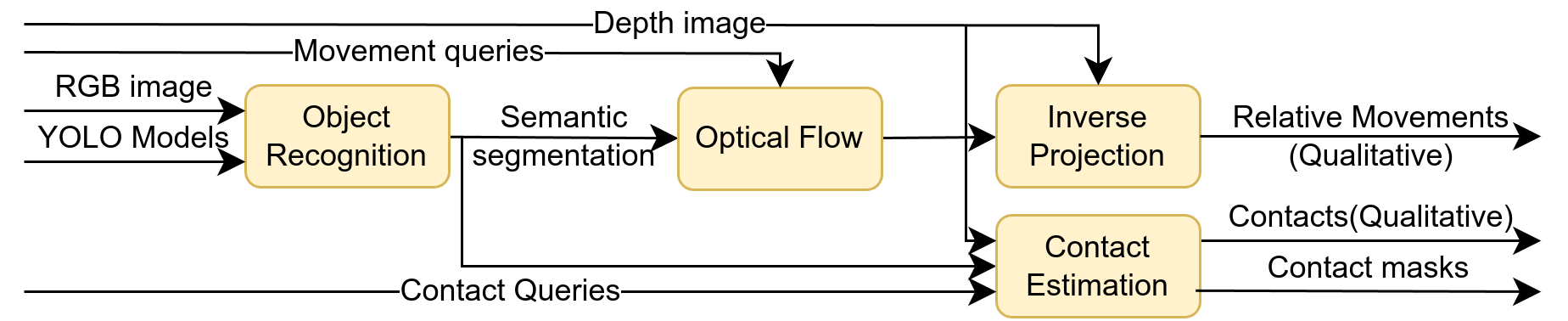}
\caption{Overview of the perception module.}
\label{fig:perception}
\end{figure}

\begin{figure}[ht]
\centering
\includegraphics[width=\columnwidth]{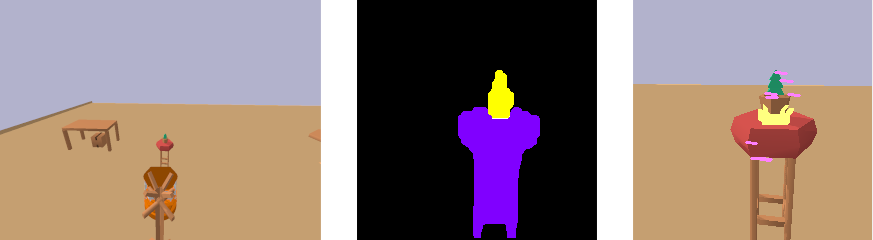}
\caption{\textit{Left:} a 3rd person view of the turtlebot in the scene. \textit{Middle:} segmentation masks from YOLO. \textit{Right:} optical flow points (purple) and contact masks (yellow) superimposed on the robot's RGB image.}
\label{fig:perception-screenshots}
\end{figure}

As shown in Fig. \ref{fig:perception},
the perception module produces annotations of pixels in the image. Neural networks -- YOLOv8 models \cite{Jocher_Ultralytics_YOLO_2023} -- are used to annotate pixels as belonging to one of several classes of interesting objects; see Fig. \ref{fig:perception-screenshots}. These annotations are referred to as ``segmentations'' (of the image into objects). Another annotator is the contact region annotator, which flags pixels close to an area where two objects of interest are in contact.

The main perception output is a set of qualitative descriptions expressed as triples of forms $(p s o)$, $(-p s o)$ where $p$ can be $contacts$, $approaches$, $departs$, $stillness$, and $-p$ can be $-contact$ (objects not in contact), and a set of contact masks. For perception to assert any triple, or produce contact masks, it has to be asked to look. Without queries, there will be no perception results. 

\subsection{The Reasoning Module}
\label{sec:reasoning_module}

\begin{figure}[ht]
\centering
\includegraphics[width=\columnwidth]{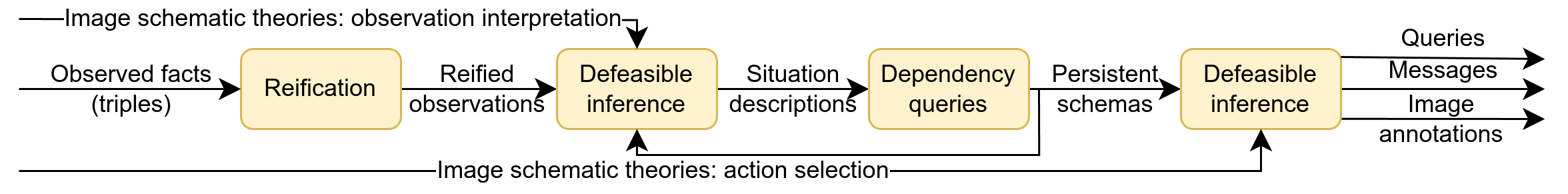}
\caption{Overview of the reasoning components.}
\label{fig:reasoning}
\end{figure}

The reasoning component is responsible for maintaining a belief state about the situation and deciding what to do based on that belief, see Fig. \ref{fig:reasoning} for an overview. Most of the reasoning is done with (defeasible) rules. For the work in this paper, defeasibility was not yet used and the inferences can be implemented via SWRL.

The main constituent of the robot's belief is a set of persistent (image) schemas, i.e. assertions about relationships between objects such as \contact, \support, as well as assertions about the robot's ``goals''. Reification steps are necessary because one can assert statements about image schematic relationships, e.g. the presence of an image schema may prevent a goal from being fulfilled. Thus, a statement coming from perception that $(contact \ a \ b)$ must be converted into a set of statements about the existence of an entity of type \emph{Contact}, with participants $a$ and $b$ -- and this \emph{Contact} entity is then further related to other image schemas perceived by the agent. Reifications also introduce new entities: the rule engine our agent uses does not allow existential rules, however the inference that certain special, ``reifiable'' predicates hold for a pair of entities will trigger the creation of a new entity and relationships to that predicate's arguments.

The reasoning module handles statements represented as triples, so these can be said to describe a graph. Some queries are convenient to handle with dedicated graph connectivity tests as opposed to rule-based inference, and thus received their own submodule: ``dependency queries'', which find entities/relationships necessary for some path to exist between certain other entities.

\section{Image-Schematic Knowledge For a Naive Theory of Support}
\label{sec:agents-theories}

We now present part of our agent's theory related to the image schema \emph{Support}.
Our agent uses a rule engine and reification in its reasoning and thus we will give the axioms here not as rules (all variables universally quantified), but as FOL expressions. Variables are expressed in bold font.


We first describe a theory of \emph{Support} situations, which we take to be that an object is supported if and only if it does not fall. We allow ourselves to make use of two categories of atypical objects -- the \emph{floor}, which exerts gravity from a distance, and \emph{Fixed} objects, unmovable by force application. Other \emph{Object}s are ``typical'': they can be moved by forces, and do not exert forces remotely. \emph{Force}s act on objects and have \emph{Direction}s, and directions may be opposite each other. For space reasons, we leave out disjointness of \emph{Fixed} and \emph{Object} and their common superclass of physical entities that can exert forces, and domain and range axioms that can be filled in by the reader from the informal glosses below.


A force that affects an object is exerted by some object (axiom \ref{axiom:affects}). Gravity is a force with direction down (ax. \ref{axiom:gravity1}), exerted by the floor, and acting on all objects (ax. \ref{axiom:gravity2}). If an object other than the floor exerts a force on another, the two are in contact (ax. \ref{axiom:contact_situation1}); if two objects are in contact, they exert forces on one another (ax. \ref{axiom:contact_situation2}). The floor can exert an upward force only when in contact (ax. \ref{axiom:floor}).

\begin{align}
\begin{aligned}
\label{axiom:affects}
\forall \mathbf{f},\mathbf{o} &: aff(\mathbf{f},\mathbf{o}) \rightarrow (\exists \mathbf{o2} : exrt(\mathbf{o2},\mathbf{f}))
\end{aligned}\\
\begin{aligned}  
\label{axiom:gravity1}
\forall \mathbf{f} &: Grv(\mathbf{f}) \rightarrow Frc(\mathbf{f}) \wedge dir(\mathbf{f},down)
\end{aligned}\\
\begin{aligned}
\label{axiom:gravity2}
\forall \mathbf{o} &: Obj(\mathbf{o}) \rightarrow (\exists \mathbf{f} : exrt(floor, \mathbf{f}) \wedge Grv(\mathbf{f}) \wedge aff(\mathbf{f}, \mathbf{o}))
\end{aligned}\\
\begin{aligned}
\label{axiom:contact_situation1}
\forall \mathbf{o1}, \mathbf{o2}, \mathbf{f} : & (\mathbf{o1} \neq floor) \wedge exrt(\mathbf{o1},\mathbf{f}) \wedge aff(\mathbf{f},\mathbf{o2}) \rightarrow \\ & (\exists \mathbf{c} : Con(\mathbf{c}) \wedge hasPrtcp(\mathbf{o1}) \wedge hasPrtcp(\mathbf{o2}))
\end{aligned}\\
\begin{aligned}
\label{axiom:contact_situation2}
    \forall \mathbf{c}, \mathbf{o1}, \mathbf{o2} : & Con(\mathbf{c}) \wedge hasPrtcp(\mathbf{o1}) \wedge hasPrtcp(\mathbf{o2}) \rightarrow \\& (\exists \mathbf{f} : exrt(\mathbf{o1},\mathbf{f}) \wedge aff(\mathbf{f},\mathbf{o2}))
\end{aligned}\\
\begin{aligned}
\label{axiom:floor}
    \forall \mathbf{f},\mathbf{o} : &exrt(floor,\mathbf{f}) \wedge aff(\mathbf{f}, \mathbf{o}) \wedge dir(\mathbf{f},up) \rightarrow \\& (\exists \mathbf{c} : Con(\mathbf{c}) \wedge hasPrtcp(\mathbf{c},floor) \wedge hasPrtcp(\mathbf{c},\mathbf{o}))
\end{aligned}
\end{align}

If a ``typical'' object does not move in the direction of a force exerted on it, then another force acts on that object in opposite direction (ax. \ref{axiom:opposite_force}). If an object exerts an upward force on another - through contact - then it is below (ax. \ref{axiom:below}).
\begin{align}
\begin{aligned}
\label{axiom:opposite_force}
\forall \mathbf{o},\mathbf{f},\mathbf{d} : & Obj(\mathbf{o}) \wedge aff(\mathbf{f},\mathbf{o}) \wedge dir(\mathbf{f},\mathbf{d}) \wedge \neg movDir(\mathbf{o},\mathbf{d}) \rightarrow \\ & (\exists \mathbf{f2},\mathbf{d2} : aff(\mathbf{f2},\mathbf{o}) \wedge dir(\mathbf{f2},\mathbf{d2}) \wedge opp(\mathbf{d2},\mathbf{d}) \\ & \neg exrt(\mathbf{o}, \mathbf{f2}))
\end{aligned}\\
\begin{aligned}
\label{axiom:below}
\forall \mathbf{o1}, \mathbf{o2}, \mathbf{f} : exrt(\mathbf{o1},\mathbf{f}) \wedge aff(\mathbf{f},\mathbf{o2}) \wedge dir(\mathbf{f},up) \rightarrow below(\mathbf{o1},\mathbf{o2})
\end{aligned}
\end{align}

``Typical'' objects only have ``typical'' parts (ax. \ref{axiom:part}). Forces exerted on/by objects are exerted on/by parts of them (ax. \ref{axiom:forced_part}, \ref{axiom:forcing_part}).

\begin{align}
\begin{aligned}
\label{axiom:part}
\forall \mathbf{o},\mathbf{p}: Obj(\mathbf{o}),hasPrt(\mathbf{o},\mathbf{p}) \rightarrow Obj(\mathbf{p})
\end{aligned}\\
\begin{aligned}
\label{axiom:forced_part}
\forall \mathbf{o},\mathbf{f} : exrt(\mathbf{o},\mathbf{f}) \rightarrow (\exists \mathbf{p} : hasPrt(\mathbf{o},\mathbf{p}) \wedge exrt(\mathbf{p},\mathbf{f}))
\end{aligned}\\
\begin{aligned}
\label{axiom:forcing_part}
\forall \mathbf{o},\mathbf{f} : aff(\mathbf{f},\mathbf{o}) \rightarrow (\exists \mathbf{p} : hasPrt(\mathbf{o},\mathbf{p}) \wedge aff(\mathbf{f},\mathbf{p}))
\end{aligned}
\end{align}

A \emph{Support} situation has supportee and supporter (ax. \ref{axiom:support_situation}). A supportee does not fall (ax. \ref{axiom:supportee}), a supporter exerts an upwards force on the supportee (ax. \ref{axiom:supporter}).

\begin{align}
\begin{aligned}
\label{axiom:support_situation}
\forall \mathbf{s} : Supp(\mathbf{s}) \rightarrow (\exists \mathbf{e}, \mathbf{r} : suppee(\mathbf{s},\mathbf{e}) \wedge supper(\mathbf{s},\mathbf{r}))
\end{aligned}\\
\begin{aligned}
\label{axiom:supportee}
\forall \mathbf{s},\mathbf{e} : suppee(\mathbf{s},\mathbf{e}) \rightarrow \neg movDir(\mathbf{e},down)
\end{aligned}\\
\begin{aligned}
\label{axiom:supporter}
\forall \mathbf{s}, \mathbf{e}, \mathbf{r} : & Supp(\mathbf{s}) \wedge suppee(\mathbf{s},\mathbf{e}) \wedge supper(\mathbf{s},\mathbf{r}) \rightarrow \\& (\exists \mathbf{f} : exrt(\mathbf{r},\mathbf{f}) \wedge aff(\mathbf{f},\mathbf{e}) \wedge dir(\mathbf{f},up))
\end{aligned}
\end{align}

Our agent actually uses descriptions of situations, so we need axioms to tell it what to query from perception to check that a Support description still applies, and upon what perceptual results it should come to believe a Support description applies. I.e., what are consequences of a Support situation in the above theory become expectations to be had if a Support description is believed to be satisfied (ax. \ref{axiom:DSupp}), and symptoms to diagnose as a Support description applying (ax. \ref{axiom:DSupp_participants}).

\begin{align}
\begin{aligned}
\label{axiom:DSupp}
     &\forall \mathbf{s},\mathbf{e} : DSupp(\mathbf{s}) \wedge suppee(\mathbf{s},\mathbf{e}) \rightarrow qrelMov(\mathbf{e},floor) \wedge qCon(\mathbf{e})
\end{aligned}\\
\begin{aligned}
\label{axiom:DSupp_participants}
    \forall \mathbf{e},\mathbf{r},\mathbf{c} : & Con(\mathbf{c}) \wedge hasPrtcp(\mathbf{c},\mathbf{e}) \wedge hasPrtcp(\mathbf{c},\mathbf{r}) \wedge below(\mathbf{r},\mathbf{e}) \wedge \\& \neg movDir(\mathbf{e},down) \rightarrow (\exists \mathbf{s} : DSupp(\mathbf{e}) \wedge suppee(\mathbf{s},\mathbf{e}) \wedge supper(\mathbf{s},\mathbf{r}))
\end{aligned}
\end{align}

The descriptions an agent believes, and the perception results they are based on/applied to, are attached to one iteration of its perception-action loop. Perception queries produced at one iteration constrain available results at the next.

\section{Functional Object Parts: Define, Recognize, Use}
\label{sec:functional-parts-define-recognize-use}

We now describe how the agent's theories described in Section~\ref{sec:agents-theories} and the perception system pick out a new object concept, which is then used to teach perception to recognize the object, and how it makes motion planning queries feasible.

Predicates such as $qCon(\mathbf{e})$ are a trigger for perception to look for objects in contact with $\mathbf{e}$. Perception returns not only a statement of two objects being in contact, but a ``contact mask'': points near where this contact occurs. The theory of support asserts parts in contact also exert and are affected by forces relevant in the support situation. For an object class, e.g. \emph{Mug}, which the agent observes supported by a \emph{Hook}, a functional part, used to support the mug using the hook, is one for which there is an observation of the part playing the supportee role:

\begin{equation}
\begin{split}
    \forall \mathbf{x}: & MugSuppbyHook(\mathbf{x}) \leftrightarrow \\&(\exists \mathbf{c}, \mathbf{s},\mathbf{m},\mathbf{h} : Con(\mathbf{c}) \wedge DSupp(\mathbf{s}) \wedge Hook(\mathbf{h}) \wedge Mug(\mathbf{m}) \wedge hasPrt(\mathbf{m},\mathbf{x})\\& \wedge suppee(\mathbf{s},\mathbf{m}) \wedge supper(\mathbf{s},\mathbf{h}) \wedge hasPrtcp(\mathbf{c},\mathbf{x})\wedge hasPrtcp(\mathbf{c},\mathbf{h}) \wedge below(\mathbf{h},\mathbf{x}))
\end{split}
\end{equation}

Treating the definition of \emph{MugSuppByHook} as a new concept allows the agent to collect images and contact masks that are observations of its instances, and retrain the neural network responsible for object detection. Thus, \emph{MugSuppByHook} becomes a class of ``object'' the network can recognize, like \emph{Mug} and \emph{Hook} are in this example. Crucially, the network can recognize a \emph{MugSuppByHook} even outside of a ``supported by \emph{Hook}'' situation. What object detection labels as \emph{MugSuppByHook} is such that it can play a supportee role in a possible situation.

\begin{figure}[ht]
\centering
\includegraphics[width=0.7\columnwidth]{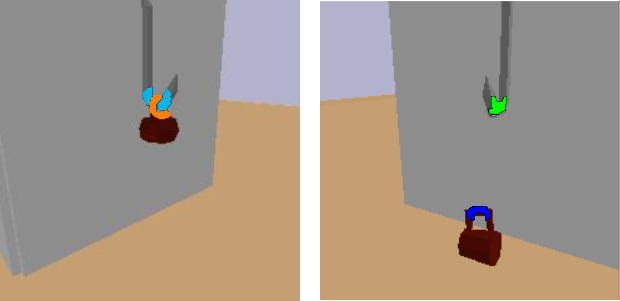}
\caption{Functional parts. Left: a frame stored for training, with annotations of functional parts. Right: frame where the newly trained network is used to recognize functional parts.}
\label{fig:fnparts-train-rec}
\end{figure}

The \emph{MugSuppByHook} object is useful when the agent is given a goal to support the mug from a hook. This requires knowing what mug target pose would prevent its falling away from the hook. The theory of support provides necessary conditions: the mug and hook must be in contact, with the mug above; the regions occupied by mug and hook should not overlap. A constraint solver can search for a satisfying pose\footnote{Since we have depth data, what the agent sees as labeled objects are sets of 3D voxels which can be moved and checked for collisions to find coordinates satisfying some symbolic constraint.}, but many of the satisfying poses will not result in the mug being supported, e.g. having just the outside bottom of the mug touch the hook is not a stable configuration and the mug will fall.

Using the part of a mug labeled as \emph{MugSuppByHook}, instead of the whole mug, as the entity for which to solve constraints 
reduces the search space, and also makes it virtually guaranteed that if a pose satisfying the constraint is found, then the mug is in fact supported by the hook.


%

\section{Discussion, Conclusions, and Future Work}
\label{sec:conclusions}

In this work we presented a neuro-symbolic architecture for cognitively inspired reactive robotics
and we escaped Dreyfus' ontology trap via learning new functional properties of existing objects and creating new concepts starting from sensor data.
We have shown how through a combination of symbolic inference driven by image schematic knowledge, and numerical procedures such as perception algorithms and geometric constraint solvers, an artificial agent is able to recognize from observed situations examples of ``functional parts'', i.e. parts of objects that are relevant for particular image schemas. Through associating the concept expression for a functional part to a set of observations it is then possible to train perception to recognize a new kind of object. Further, because the new object is a functional part it assists in the solution of motion planning tasks, because it focuses the search for arrangements conducive to manifesting an image schema.

We are pursuing several avenues for continuation. One is to expand the role of geometric inference and physics simulation so as to incorporate anticipation. The main direction however is to expand the depth of time that the agent can consider. Its current operation attends only to the current moment, with the previous visual image used to compute relative movements. Schemas are persistent only in the sense that a set of triples that held at a previous step may still hold now, and previously captured images with annotations of functional parts are treated as independent of each other. However, functional parts involved in a situation often become obscured from vision by performing their role; e.g. the container part of a spoon sinking into a soup. Thus, it is necessary to annotate functional parts on frames where they do not yet perform the function, which requires, at the symbolic level, an understanding of a sequence of frames as observing a process with image-schematic consequences, and at the numeric level techniques to match frame parts assumed to exist at a particular location but invisible.

\section*{Acknowledgements}
This work was supported by the Future Artificial Intelligence Research (FAIR) project, code PE00000013 CUP 53C22003630006, the German Research Foundation DFG, as part of Collaborative Research Center (Sonderforschungsbereich) 1320 Project-ID 329551904 “EASE - Everyday Activity Science and Engineering”, subproject ``P01 -- Embodied semantics for the language of action and change: Combining analysis, reasoning and simulation'', and by the
FET-Open Project \#951846 ``MUHAI – Meaning and Understanding for Human-centric AI'' by the EU Pathfinder and Horizon 2020 Program.


\bibliographystyle{vancouver}
\bibliography{fois}

\end{document}